\pdfoutput=1

\documentclass[11pt]{article}

\usepackage[]{acl}

\usepackage{times}
\usepackage{latexsym}

\usepackage[T1]{fontenc}

\usepackage[utf8]{inputenc}

\usepackage{microtype}

\usepackage{inconsolata}

\usepackage{amssymb, amsmath}
\usepackage{siunitx}
\usepackage{bm}
\usepackage{graphicx}
\usepackage{booktabs}
\usepackage{tabularx}
\usepackage{multirow}
\usepackage{adjustbox}
\usepackage{array}
\usepackage{pifont}
\usepackage{xspace}
\usepackage{tcolorbox}
\usepackage{colortbl}
\usepackage{hyperref}
\usepackage{arydshln}
\usepackage{wrapfig}
\usepackage{soul}
\usepackage{makecell}
\usepackage{color, xcolor}
\usepackage{pgfplots}
\definecolor{tiffanyblue}{RGB}{129,216,208}
\definecolor{bangdiblue}{RGB}{0,149,182}
\definecolor{kleinblue}{RGB}{0,47,167}
\usepackage{amssymb}
\usepackage{circledtext}
\usetikzlibrary{shapes}
\usetikzlibrary{arrows,decorations.pathmorphing,backgrounds,positioning,fit,petri}
\usetikzlibrary{arrows.meta,fit,shapes.arrows}
\usetikzlibrary{positioning,shadows,patterns}
\usepackage{caption} 
\usepackage{tikz}
\usepackage{subfig}
\usepackage{tkz-kiviat,pgfplots}
\pgfplotsset{compat=newest}

\newcommand{\ours}{\textsc{Start}\xspace}
\newcommand{\headercolor}{\rowcolor{gray!15}}

\newcommand{\eg}{\emph{e.g.}}
\makeatletter
\def\adl@drawiv#1#2#3{%
        \hskip.5\tabcolsep
        \xleaders#3{#2.5\@tempdimb #1{1}#2.5\@tempdimb}%
                #2\z@ plus1fil minus1fil\relax
        \hskip.5\tabcolsep}
\newcommand{\cdashlinelr}[1]{%
  \noalign{\vskip\aboverulesep
           \global\let\@dashdrawstore\adl@draw
           \global\let\adl@draw\adl@drawiv}
  \cdashline{#1}
  \noalign{\global\let\adl@draw\@dashdrawstore
           \vskip\belowrulesep}}
\makeatother

\title{Advancing Large Language Model Attribution through Self-Improving}
\author{
 \textbf{Lei Huang\textsuperscript{1}},
 \textbf{Xiaocheng Feng\textsuperscript{1,2}\thanks{Corresponding Author}},
 \textbf{Weitao Ma\textsuperscript{1}},
 \textbf{Liang Zhao\textsuperscript{1}},
 \textbf{Yuchun Fan\textsuperscript{3}},
\\
 \textbf{Weihong Zhong\textsuperscript{1}},
 \textbf{Dongliang Xu\textsuperscript{4}},
 \textbf{Qing Yang\textsuperscript{4}},
 \textbf{Hongtao Liu\textsuperscript{4}},
 \textbf{Bing Qin\textsuperscript{1,2}}
\\
 \textsuperscript{1} Harbin Institute of Technology, Harbin, China 
 \\
 \textsuperscript{2} Peng Cheng Laboratory, Shenzhen, China \quad \quad \textsuperscript{3} Northeastern University, Shenyang, China
 \\
 \textsuperscript{4} Du Xiaoman Science Technology Co., Ltd., Beijing, China
 \\
 \texttt{\{lhuang, xcfeng, wtma, lzhao, whzhong, qinb\}@ir.hit.edu.cn}
 \\
 \texttt{yuchunfan\_neu@outlook.com}
 \\
 \texttt{\{xudongliang, yangqing, liuhongtao01\}@duxiaoman.com}
}

\begin{document}
\maketitle

\begin{abstract}
Teaching large language models (LLMs) to generate text with citations to evidence sources can mitigate hallucinations and enhance verifiability in information-seeking systems. However, improving this capability requires high-quality attribution data, which is costly and labor-intensive. 
Inspired by recent advances in self-improvement that enhance LLMs without manual annotation, we present \ours, a \textbf{S}elf-\textbf{T}aught \textbf{A}tt\textbf{R}ibu\textbf{T}ion framework for iteratively improving the attribution capability of LLMs.
First, to prevent models from stagnating due to initially insufficient supervision signals, \ours leverages the model to self-construct synthetic training data for warming up.
To further improve the model's attribution ability, \ours iteratively utilizes fine-grained preference supervision signals constructed from its sampled responses to encourage robust, comprehensive, and attributable generation.
Experiments on three open-domain question-answering datasets, covering long-form QA and multi-step reasoning, demonstrate significant performance gains of 25.13\% on average without relying on human annotations and more advanced models.
Further analysis reveals that \ours excels in aggregating information across multiple sources.
\end{abstract}
\section{Introduction}
The rapid development of large language models (LLMs) \citep{openai2023gpt4, zhao2023a} has led to their prosperity as indispensable tools for information seeking. Despite their remarkable capability to generate fluent and informative responses to user queries, LLMs also struggle with hallucinations \citep{huang2023survey}. 
To facilitate factuality verification, recent research \citep{bohnet2022attributed} has explored attributed text generation, a paradigm that enables LLMs to generate responses with citations. By attributing models' output to verifiable sources, it can improve the explainability and credibility of LLM-generated content \citep{li2023a}.

While beneficial, the ability to attribute contextual sources is not inherent in LLMs. Most work induces LLMs to generate text with citations via in-context learning \citep{gao2023enabling}, which is far from satisfactory \citep{liu2023evaluating}.
The current winning recipe for accurate attribution involves fine-tuning on high-quality attribution responses\footnote{\textit{Attribution responses} refers to ``responses with in-line citations, \eg, [1][2]''.}~\citep{li2024improving}. 
However, acquiring such data typically requires either manual curation \citep{malaviya2023expertqa}, or distilled from the most advanced LLMs \citep{huang2024training, huang2024learning}, both of which are costly and not scalable, thus limiting the growth of models' attribution capability.
One promising solution is self-improvement \citep{yuan2023scaling}, which has demonstrated the potential to boost model performance by learning from self-generated high-quality samples.

Inspired by this, we aim to explore the potential of self-improvement in bootstrapping the attribution ability of LLMs. However, achieving this goal presents several challenges. 
One significant challenge lies in the risk of \textit{model stagnation} during the self-improvement process, primarily due to the insufficient supervision signals obtained in the early stage.
Concretely, considering the inferior performance of LLMs in handling the attribution task \citep{gao2023enabling}, generating sufficient high-quality attribution responses solely through sampling proves difficult. This scarcity of high-quality samples limits the opportunities for LLMs to self-improve effectively.
Another challenge stems from the limitation of \textit{weak supervision signals}.
Current self-improvement approaches \citep{yuan2023scaling} primarily involve supervised fine-tuning on high-quality samples while discarding low-quality ones. When applied to LLM attribution, these high-quality samples provide only weak supervision signals, mainly teaching LLMs on the surface form of attribution (\eg, proper citation format) \citep{li2024improving}.
Such practice may neglect the potential of exploring fine-grained signals from low-quality samples to learn what constitutes a desirable attribution response.

To address these challenges, we present \ours, a \textbf{S}elf-\textbf{T}aught \textbf{A}tt\textbf{R}ibu\textbf{T}ion framework designed to bootstrap the attribution capabilities of LLMs.
To prevent models from stagnating early due to insufficient supervision signals, we first leverage the model to self-construct high-quality synthetic attribution data (\S\ref{sec:data_synthesize}). 
The data synthesis process follows \textit{reverse attribution thinking}: the model initially generates a response to a given query, then breaks it into atomic claims, and finally randomly combines them to create synthetic documents. This process not only simulates multi-source information-seeking scenarios but also ensures precise attribution, as each document can be directly traced back to the specific claim it originated from.
These high-quality synthetic data are then utilized for warming up, providing a good starting point for LLMs to self-improve.
Furthermore, to better explore fine-grained supervision signals for LLM attribution, we introduce an iterative self-improving recipe (\S\ref{sec:self_improving}). 
Specifically, the framework meticulously designs fine-grained rewards tailored for LLM attribution, covering robustness, comprehensiveness, and attributability. By scoring multiple candidates through sampling and selecting those with the highest holistic rewards for supervised fine-tuning, the framework subsequently utilizes low-quality samples to construct fine-grained preference pairs with diverse optimization rewards for preference optimization. This iterative process further fosters the self-improvement of attribution capabilities.

We conduct extensive experiments across three open-domain question-answering datasets, covering long-form QA and multi-step reasoning. Results indicate that \ours achieves significant performance gains of 25.13\% on average in citation quality.
Moreover, \ours successfully achieves self-improvement in LLM attribution, showing progressive improvements across iterations. 
Ablation studies confirm that each component significantly contributes to the improvement.
Further analysis shows that \ours not only excels in generating superior attributable responses but also in effectively aggregating information across multiple sources. 
\section{Related Work}

\subsection{Large Language Model Attribution}
Attribution has gained significant attention for enhancing the interpretability and verifiability of LLMs \citep{gao2023enabling, li2023a}.
Recent studies have focused on improving LLM attribution in a supervised way.
\citet{asai2023self} first distill GPT-4 to collect high-quality attribution data, aiming to teach the model to generate grounded answers with citations through self-reflecting.
Similarly, \citet{huang2024training} develop a training framework starting with distilling ChatGPT, followed by designing reward models to teach the LLM to generate highly supportive and relevant citations.
Additionally, \citet{li2024improving} model the attribution task from a preference learning perspective, where they first fine-tune the model on human-labeled attribution datasets and then perform preference optimization using synthesized preference data.
Furthermore, \citet{huang2024learning} take this further by extending the attribution format to a fine-grained citation level, primarily distilled from ChatGPT. It enables the model to first ground the fine-grained quotes within the context and then condition the generation process on them.
In contrast to these methods, \ours aims to bootstrap attribution capability without relying on human-labeled data or distilling from more capable LLMs. 
\begin{figure*}[h]
    \centering
    \includegraphics[width=1.0\textwidth]{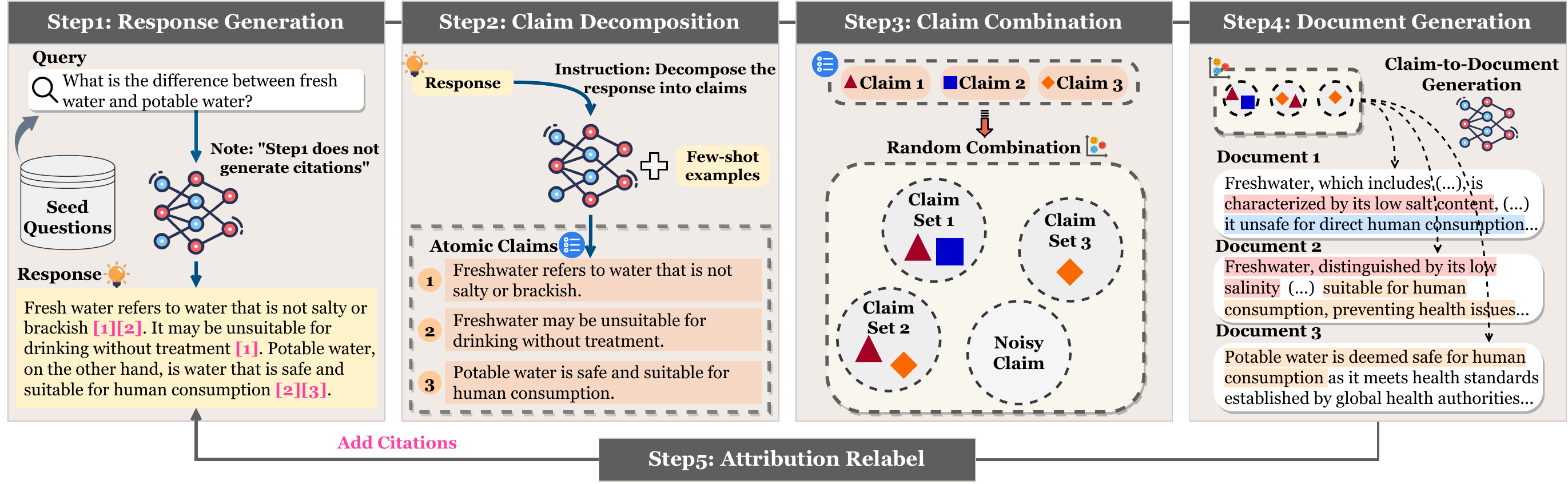}
    \caption{The data synthesis pipeline consists of five steps: given a user query, the LLM first generates an informative response \textbf{without citations} in a closed-book setting. Subsequently, the LLM decomposes this response into atomic claims. These claims are then \textbf{randomly} grouped into specific sets, which serve as the basis for generating documents that cover all included claims. Finally, we trace back to the initial response to relabel the citations.}
    \label{fig:data-synthesize}
\end{figure*}
\subsection{Self-Improvement for LLMs}
High-quality data either human-crafted or distilled from advanced LLMs has proven effective in enhancing the performance of LLMs.
However, acquiring such high-quality data can be prohibitively expensive. 
Recently, self-improvement approaches \citep{gulcehre2023reinforced, yuan2024self}, where LLMs learn from self-generated samples have emerged as a viable solution to compensate for the scarcity of high-quality data. 
These methods typically involve employing heuristic rules \citep{zelikman2022star}, self-critique \citep{tian2024toward}, or training additional verifiers \citep{hosseini2024vstar} to assess the quality of model-generated samples.
Such practices are particularly effective in reasoning tasks, \eg, mathematical reasoning, where LLMs already demonstrate capable abilities and can receive precise feedback on correctness.
However, these advantages are absent in the attribution task, due to its challenging nature.
To bridge the gap, we take an initial step towards exploring the potential of self-improvement in LLM attribution.

\section{Problem Formulation and Methodology}
\label{sec:methods}
We follow a formulation of attributed text generation as described in \citet{gao2023enabling}. This task involves processing a user query $q$ for information-seeking, given a corpus of retrieved documents $\mathcal{D}$, to generate a response $\mathcal{S}$ with in-line citations.
We assume the response $\mathcal{S}$ as consisting of $n$ statements, such that $\mathcal{S} = \{s_1, s_2, \ldots, s_n \}$. Each statement $s_i \in \mathcal{S}$ cites a list of passage $\mathcal{C}_i = \{c_{i1}, c_{i2}, \ldots\}$, where $c_{ij} \in \mathcal{D}$.
Citations are presented in the form of \texttt{[1][2]}, which represent the attribution to specific documents in $\mathcal{D}$.

\begin{figure*}[h]
    \centering
    \includegraphics[width=1.0\textwidth]{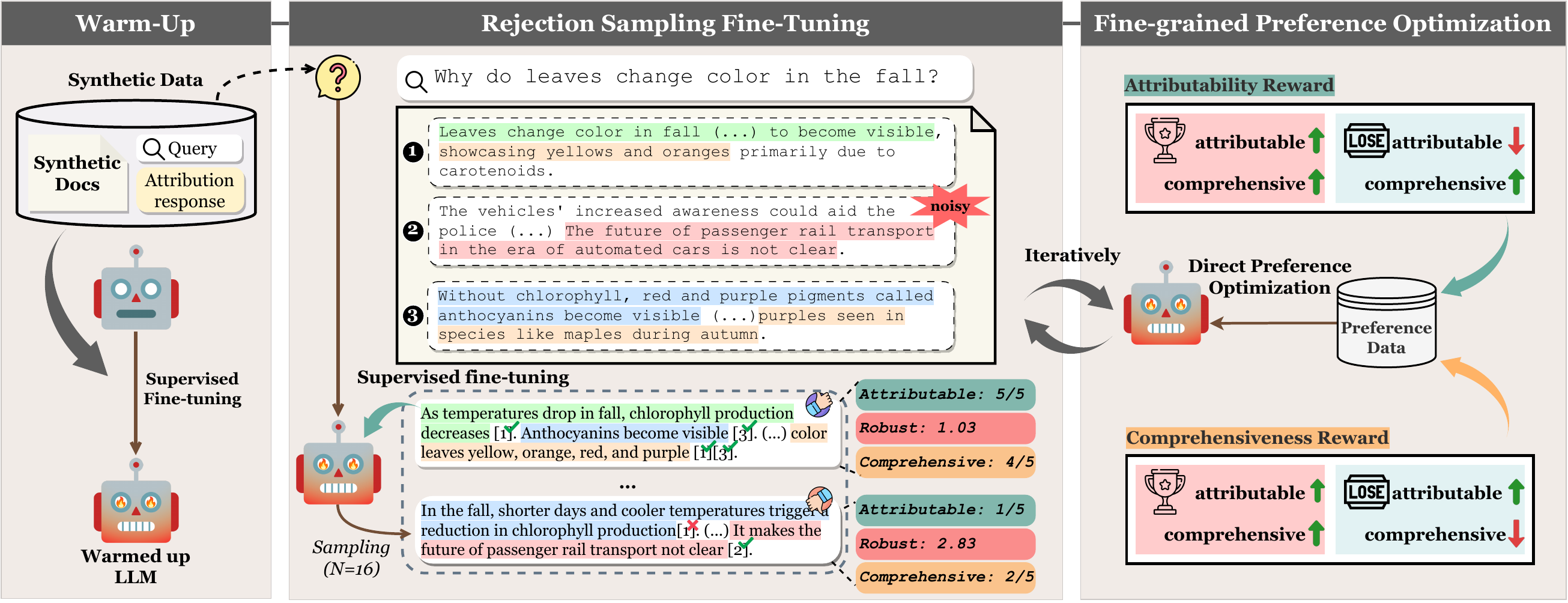}
    \caption{Overview of our self-improving framework, which consists of two stages. The model is first warmed up using synthetic data (\S\ref{sec:data_synthesize}). This provides a good starting point to enable the model to generate high-quality samples in the subsequent iterative training. Next, the model is further trained via rejection sampling fine-tuning and fine-grained preference optimization iteratively (\S\ref{sec:self_improving}). This iterative process bootstraps the model's attribution capability by fully utilizing the supervision signals from its sampled generations.}
    \label{fig:training-framework}
\end{figure*}

Next, we present an overview of \ours, a training framework designed to teach LLMs to self-improve their attribution ability, as illustrated in Figure~\ref{fig:training-framework}. \ours consists of two essential stages: synthetic data warm-up (\S\ref{sec:data_synthesize}) and self-improving for LLM attribution (\S\ref{sec:self_improving}).

\subsection{Synthetic Data Warm-Up}
\label{sec:data_synthesize}
The core of self-improvement lies in generating \textit{high-quality} samples and iteratively learning from them. Intuitively, a \textit{high-quality} attribution response should not be distracted by irrelevant documents (\textit{robustness}) and capture high coverage of viewpoints across multiple documents (\textit{comprehensiveness}) while maintaining high citation quality (\textit{attributability}). However, existing LLMs typically show inferior performance in the attribution task, significantly hindering their ability to generate such high-quality samples. This limitation poses substantial challenges to enhancing their attribution capabilities through self-improvement.

In this stage, we propose utilizing the model to self-construct high-quality synthetic data for warming up, enabling the model to have the basic ability to generate robust, comprehensive, and attributable responses across multiple sources. The pipeline consists of the following steps, shown in Figure~\ref{fig:data-synthesize}. More details can be found in Appendix~\ref{appendix:data_synthesis}.

\paragraph{Step 1: Response Generation} Given an arbitrary model, we first sample a query $q$ from seed questions $Q$ and then generate a long-form answer $S$ utilizing the parametric knowledge of the model itself. The model is required to produce informative answers that cover multiple perspectives.

\paragraph{Step 2: Claim Decomposition}
Prior work \citep{min2023factscore} has explored using atomic claims as a fundamental unit in long-form text generation. Thus, for the response $S$, we ask the model to decompose it into atomic claims. Each atomic claim represents a distinct piece of information.

\paragraph{Step 3: Claim Combination} To ensure that the response behaves as an aggregation of information from multiple documents, we randomly combine different claims into one claim set. This process helps simulate the natural diversity of viewpoints and sources, thus enhancing the comprehensiveness and realism of the synthesized responses.

\paragraph{Step 4: Document Generation} For each claim set, we prompt the model to generate a synthetic document $D$ that provides a comprehensive discussion of the grouped claims. Additionally, to enhance the robustness of the response, we introduce irrelevant documents by uniformly sampling documents generated from other queries.

\paragraph{Step 5: Attribution Relabel} The final step involves labeling the response with citations from the generated documents. This process ensures that each claim within the response is explicitly attributed to its source. In this way, for each query $q$, and documents set $D$, we can obtain an informative and attributable response while maintaining robustness against irrelevant documents.

Next, the model is fine-tuned for warming up with the MLE objective on the synthesized dataset, which consists of $N$ data entries, each containing a query $q_i$, a document set $\mathcal{D}_i$, and a high-quality attributable response $y_i$:

\begin{equation}
    \mathcal{L} = -\sum_{i=1}^{N} \log P(y_i | q_i, \mathcal{D}_i; \theta)
\end{equation}
\subsection{Self-Improving for LLM Attribution}
\label{sec:self_improving}
In this stage, we propose to iteratively boost the model's attribution capability by exploring more fine-grained supervision signals, rather than solely relying on \textit{golden} responses in synthetic data. This involves leveraging rejection sampling for \textit{data growing} and fine-grained preference optimization for \textit{capability evolution}.\subsubsection{Rejection Sampling Fine-tuning}
After warming up, we first sample $N$ candidates for each query in the synthetic dataset and then score each candidate with fine-grained rewards that cover three key dimensions: \textit{robustness}, \textit{comprehensiveness}, and \textit{attributability}.
\paragraph{Attributability} serves as the indispensable condition for high-quality attributable generation. It quantifies the extent to which a response is fully supported by the cited documents. To accurately measure attributability, we employ an off-the-shelf Natural Language Inference (NLI) model\footnote{\url{huggingface.co/google/t5_xxl_true_nli_mixture}} by checking whether each statement in the response is entailed by the corresponding cited documents.
\begin{equation}
    \text{AttrScore} = \frac{1}{S} \sum_{i=1}^{S} \texttt{Entail}(\text{Docs}, \text{statement}_i)
\end{equation}
where $S$ is the total number of statements in the response and \texttt{Entail} returns 1 if the statement $i$ is entailed by cited documents, and 0 otherwise.
\paragraph{Robustness} measures the degree to which a model-generated response is influenced by irrelevant contexts. Considering that we can identify relevant documents $d_r$ within the document set $D$ for each query $q$, thus we quantify robustness by calculating the probability difference of the model $M$ to generate the response $y$ under different contexts. The robustness score is defined as follows:
\begin{equation}
    \text{RobustScore} = \frac{P_M(y \mid q \oplus d_r)}{P_M(y \mid q \oplus D)}
\end{equation}
Empirically, the closer the score is to 1, the less the response is disturbed by irrelevant documents.

\paragraph{Comprehensiveness} measures the extent to which a response captures all relevant information from the source documents. As the \textit{golden} responses in the synthetic data are designed to aggregate and reflect information across multi-documents, thus we quantify comprehensiveness by decomposing them into sub-claims and verifying whether these claims are covered by the sampled generation $y$.
 We compute the score as below:
\begin{equation}
    \text{CompreScore} = \frac{1}{C} \sum_{i=1}^{C} \texttt{Entail}(\text{claim}_i, y)
\end{equation}
where $\text{claim}_i$ represents sub-claims and $C$ is the number of \textit{golden} sub-claims.

Subsequently, we formulate a holistic reward function (Eq.~\ref{eq:reward}) considering the above dimensions. This function is employed to rank generated candidates, with the top-ranked candidate being selected for further supervised fine-tuning.
\begin{equation}
    \text{Reward} = \mathbb{I}(\text{AttrScore}) \times \frac{\text{CompreScore}}{\text{RobustScore}}
    \label{eq:reward}
\end{equation}
Here, $\mathbb{I}$ is an indicator function that returns 1 if $\text{AttrScore}=1$, and 0 otherwise.
\subsubsection{Fine-grained Preference Optimization}
The common way of self-improvement focuses on updating the model with high-quality samples while discarding low-quality ones. 
For LLM attribution, simply supervised fine-tuning with highly attributable responses only teaches the LLM to learn surface characteristics of attribution, \eg, the correct form of citation.
Inspired by human cognition, learning from mistakes provides more fine-grained signals to understand the mechanisms that drive successful attribution than simply imitating correct examples.
Thus, we aim to fully unlock the potential of low-quality samples by constructing fine-grained preference pairs with different optimization rewards for preference optimization.

Given the multi-objective nature of LLM attribution, our focus is specifically on \textit{attributability} and \textit{comprehensiveness}, utilizing corresponding rewards functions to construct preference data respectively\footnote{We do not optimize separately for robustness as the model already shows sufficient robustness after rejection sampling fine-tuning.}.
Specifically, we pair samples that exhibit high attributability but low comprehensiveness with the top-ranked sample selected using a holistic reward, and vice versa.
These preference pairs, each addressing different optimization objectives, are then aggregated to further train the LLM via DPO \citep{rafailov2023direct}:
$$\mathcal{L}_\mathrm{DPO} = -\mathbb{E}[\log \sigma (\hat{r}_\theta(x, y^+) - \hat{r}_\theta(x, y^-))]$$
\begin{equation}\label{eq:2}
\hat{r}_\theta(x, y) = \beta \log \frac{\pi_{\theta}(y\mid x)}{\pi_{\mathrm{ref}}(y\mid x)}
\end{equation}

Here, reference model $\pi_{\mathrm{ref}}$ is initialized with the model after rejection sampling to minimize the distribution shift from the reference distribution.
\section{Experiments}
\label{sec:experiments}

\subsection{Datasets}
Following previous work \citep{ye2023effective, li2024improving}, we conduct our experiments using two long-form question-answering datasets: ASQA \citep{stelmakh2022asqa} and ELI5 \citep{fan2019eli5}, as well as a multi-step reasoning dataset, StrategyQA \citep{geva2021did}. 
Both ASQA and ELI5 feature factoid long-form answers that require synthesizing highly relevant documents in response to a user query. In StrategyQA, answers demand a combination of information-seeking and implicit reasoning.
Further details on the data statistics, knowledge corpus used for retrieval, and examples for each dataset are provided in Appendix \ref{appendix:datasets}.
\subsection{Evaluation}
\definecolor{steelbluev2}{HTML}{DAE8FC}
\definecolor{steelblue}{HTML}{82B0D2}
\begin{table*}[t]
\small
  \centering
  \resizebox{\linewidth}{!}{
    \begin{tabular}{lcccccccccccc}
    \toprule
    \multirow{2}[4]{*}{\textbf{Model}} & \multicolumn{4}{c}{\textbf{ASQA}} & \multicolumn{4}{c}{\textbf{ELI5}} & \multicolumn{4}{c}{\textbf{StrategyQA}}\\
    \cmidrule(lr){2-5}
    \cmidrule(lr){6-9}
    \cmidrule(lr){10-13}
    &  \multicolumn{1}{c}{\textbf{Correctness}} & \multicolumn{3}{c}{\textbf{Citation}}  & \multicolumn{1}{c}{\textbf{Correctness}} & \multicolumn{3}{c}{\textbf{Citation}} &  \multicolumn{1}{c}{\textbf{Correctness}} &  \multicolumn{3}{c}{\textbf{Citation}} \\
    \cmidrule(lr){2-2}
    \cmidrule(lr){3-5}
    \cmidrule(lr){6-6}
    \cmidrule(lr){7-9}
    \cmidrule(lr){10-10}
    \cmidrule(lr){11-13}
    & \multicolumn{1}{c}{EM Rec.} & \multicolumn{1}{c}{Rec.} & \multicolumn{1}{c}{Prec.} & \multicolumn{1}{c}{F1.} & \multicolumn{1}{c}{Claim} & \multicolumn{1}{c}{Rec.} & \multicolumn{1}{c}{Prec.} & \multicolumn{1}{c}{F1} & \multicolumn{1}{c}{Acc.} & \multicolumn{1}{c}{Rec.} & \multicolumn{1}{c}{Prec.} & \multicolumn{1}{c}{F1} \\
    \midrule
    \headercolor
    \multicolumn{13}{c}{\textbf{\textit{In-context Learning}} \& \textbf{\textit{Post-hoc}}} \\
    \midrule
     Llama-2-13B (ICL) & 35.2 & 38.4 & 39.4 & 38.9 & 13.4 & 17.3 & 15.8 & 16.5 & 65.6  & 20.6 & 33.1 & 25.4 \\
     Llama-2-13B (PostAttr) & 25.0 & 23.6 & 23.6 & 23.6 & 7.1 & 5.7 & 5.8 & 5.8 & 64.3  & 8.7 & 8.7 & 8.7 \\
    \midrule
    \headercolor
    \multicolumn{13}{c}{\textbf{\textit{Training-based}}} \\
    \midrule
    Distill-Llama-3-70B-Instruct & 41.1 & 60.4 & 53.8 & 56.9 & 12.9 & 28.7 & 25.2 & 26.8 & 70.8  & 28.4 & 30.7 & 29.5 \\
    Distill-Mixtral-8x7B-Instruct & 40.3 & 64.9 & 63.5 & 64.2 & 13.8 & 34.3 & 35.0 & 34.6 & 63.9  & 38.4 & 49.2 & 43.1 \\
    \midrule
    Self-RAG \citep{asai2023self} & 31.7 & 70.3 & 71.3 & 70.8 & 10.7 & 20.8 & 22.5 & 21.6 & 62.1  & 31.4 & 36.5 & 33.8 \\
    AGREE \citep{ye2023effective} & 39.4 & 64.0 & 66.8 & 65.4 & 9.4 & 21.6 & 16.0 & 18.4 & 64.6  & 30.2 & 37.2 & 33.3 \\
    APO \citep{li2024improving} & 40.5 & 72.8 & 69.6 & 71.2 & \textbf{13.5} & 26.0 & 24.5 & 25.2 & 61.8  & 40.0 & 39.1 & 39.6 \\
    FGR \citep{huang2024training} & 38.7 & 73.5 & 74.7 & 74.1 & 9.8 & 53.1 & 55.9 & 54.5 & 64.9  & 29.5 & 42.4 & 34.8 \\
	\midrule
    \ours (Warming-up) & 39.2 & 23.2 & 23.9 & 23.5 & 11.9 & 9.9 & 10.2 & 10.0 & 61.2  & 9.4 & 9.6 & 9.5 \\
    \rowcolor{steelblue!33}\ours (Iteration 1) & 42.2 & 68.8 & 75.6 & 72.0 & 11.3 & 47.4 & 50.5 & 48.9 & \textbf{73.4}  & 44.4 & 48.6 & 46.4 \\
    \rowcolor{steelblue!33}\ours (Iteration 2) & 42.9 & 76.1 & 81.0 & 78.5 & 10.0 & \textbf{65.6} & 65.1 & 65.3 & 72.7  & 51.9 & 54.1 & 53.0  \\
    \rowcolor{steelblue!33}\ours (Iteration 3) & \textbf{44.2} & \textbf{76.2} & \textbf{84.2} & \textbf{80.0} & 9.6 & 62.4 & \textbf{69.1} & \textbf{65.6} & 69.6  & \textbf{60.0} & \textbf{56.6} & \textbf{58.2} \\
    \bottomrule
    \end{tabular}
  }
  \caption{Main result between our method and baselines. Experiments are evaluated on ASQA, ELI5, and StrategyQA datasets. For most baselines, we use the result of previous works \citep{asai2023self, ye2023effective, li2024improving}.}
  \label{tab:main_result}
\end{table*}%

Following previous research \citep{gao2023enabling}, we evaluate model-generated responses mainly on two dimensions: \textbf{Citation Quality} and \textbf{Correctness}. Our evaluation methodology combines both automated metrics and human evaluation.

\paragraph{Automatic Evaluation.}
To assess citation quality, we calculate the \textit{citation precision}, \textit{citation recall}, and its harmonic mean \textit{citation F1} based on the definition in \citet{gao2023enabling}. We use TRUE \citep{honovich2022true}, a T5-11B model fine-tuned on a collection of natural language inference (NLI) datasets to examine whether the cited documents entail the generated statement. 
For correctness, different datasets are measured differently. For ASQA, we report the exact match recall (\textbf{EM Rec.}) of correct short answers. For ELI5, we report the claim recall (\textbf{Claim}) by checking whether the model output entails the sub-claims generated by \texttt{text-davinci-003}. 
For StrategyQA, the format of answers begins with yes/no, we evaluate correctness by reporting the accuracy (\textbf{Acc.}). See Appendix \ref{appendix:automatic_evaluation} for more details.

\paragraph{Human Evaluation.}
We collected a total of 150 instances from the test sets of ASQA, ELI5, and StrategyQA for human evaluation, with each dataset providing 10 instances from five different systems.
The evaluation is divided into two parts: citation quality and overall quality (comprehensiveness and correctness).  
More details in Appendix~\ref{appendix:human_evaluation}.
\subsection{Baselines}
We compare \ours with the following baselines.
For more details, please refer to Appendix \ref{appendix:baselines}.
\paragraph{In-context Learning (ICL).} Following \citet{gao2023enabling}, we enable the LLM to generate citations via in-context learning. For each query, we first retrieve five relevant documents and then prompt the LLM with two-shot demonstrations.
\paragraph{Post-hoc Attribution (PostAttr).} Following \citet{ye2023effective}, given a query, we first instruct the LLM to generate an initial response leveraging its parametric knowledge. For each statement in the response, we use the NLI model\footnote{We use the same NLI model during citation evaluation.} to find the maximally supported document and cite accordingly.
\paragraph{Training-based Methods.} Training on high-quality data serves as a strong baseline to unlock the attribution ability of LLMs. We consider the following training-based methods.

\textbf{Knowledge Distillation} employs the most capable LLMs, \eg, Llama-3-70B-Instruct and Mixtral-8x7B-Instruct, as teacher models to train a student model on distilled attribution data.

\textbf{Self-RAG} \citep{asai2023self} first collect \textit{data distilled from GPT-4}, then teach the LLM to retrieve on-demand while reflecting on its generation to improve both generation quality and attributions.

\textbf{AGREE} \citep{ye2023effective} trains the LLM to self-ground its response in retrieved documents using automatically collected data and then leverages test-time adaptation to reinforce unverified statements.

\textbf{APO} \citep{li2024improving} models LLM attribution as a preference learning task, where they first supervised-fine-tuned on \textit{human-labeled high-quality data} and then automatically collect preference data for preference optimization.

\textbf{FGR} \citep{huang2024training} first collects \textit{attribution data distilled from ChatGPT} and then designs rewards tailored for LLM attribution to teach the LLM to generate supportive and relevant citations.
\subsection{Implementation Details}
For a fair comparison, all training-based baselines and \ours employ \texttt{Llama-2-13b-base}~\citep{touvron2023llama2}.  
Further details on the implementation of \ours are presented in Appendix \ref{appendix:implement}.
\section{Results}
\label{sec:results}
\subsection{Main Results}
\definecolor{steelbluev2}{HTML}{DAE8FC}
\definecolor{steelblue}{HTML}{82B0D2}
\begin{table*}[t]
\small
  \centering
  \resizebox{\linewidth}{!}{
    \begin{tabular}{lcccccccccccc}
    \toprule
    \multirow{2}[4]{*}{\textbf{Model}} & \multicolumn{4}{c}{\textbf{ASQA}} & \multicolumn{4}{c}{\textbf{ELI5}} & \multicolumn{4}{c}{\textbf{StrategyQA}}\\
    \cmidrule(lr){2-5}
    \cmidrule(lr){6-9}
    \cmidrule(lr){10-13}
    &  \multicolumn{1}{c}{\textbf{Correctness}} & \multicolumn{3}{c}{\textbf{Citation}}  & \multicolumn{1}{c}{\textbf{Correctness}} & \multicolumn{3}{c}{\textbf{Citation}} &  \multicolumn{1}{c}{\textbf{Correctness}} &  \multicolumn{3}{c}{\textbf{Citation}} \\
    \cmidrule(lr){2-2}
    \cmidrule(lr){3-5}
    \cmidrule(lr){6-6}
    \cmidrule(lr){7-9}
    \cmidrule(lr){10-10}
    \cmidrule(lr){11-13}
    & \multicolumn{1}{c}{EM Rec.} & \multicolumn{1}{c}{Rec.} & \multicolumn{1}{c}{Prec.} & \multicolumn{1}{c}{F1.} & \multicolumn{1}{c}{Claim} & \multicolumn{1}{c}{Rec.} & \multicolumn{1}{c}{Prec.} & \multicolumn{1}{c}{F1} & \multicolumn{1}{c}{Acc.} & \multicolumn{1}{c}{Rec.} & \multicolumn{1}{c}{Prec.} & \multicolumn{1}{c}{F1} \\
    \midrule
    \rowcolor{gray!10}\ours (Iteration 1) & 42.2 & 68.8 & 75.6 & 72.0 & 11.3 & 47.4 & 50.5 & 48.9 & 73.4 & 44.4 & 48.6 & 46.4 \\
    \ \ w/o. warm-up & 35.7 & 36.3 & 32.7 & 34.4 & 12.1 & 15.2 & 13.7 & 14.4 & 65.9  & 18.0 & 17.2 & 17.6 \\
    \ \ w/o. preference & 40.6 & 42.2 & 47.2 & 44.6 & 12.9 & 16.5 & 17.4 & 16.9 & 63.7  & 21.5 & 24.6 & 22.9 \\
    \cdashlinelr{1-13}
    \rowcolor{gray!10}\ours (Iteration 2) & 42.9 & 76.1 & 81.0 & 78.5 & 10.0 & 65.6 & 65.1 & 65.3 & 72.7  & 51.9 & 54.1 & 53.0  \\
    \ \ w/o. warm-up & 33.5 & 57.4 & 52.1 & 54.6 & 10.0 & 26.7 & 23.0 & 24.7 & 69.0  & 32.4 & 33.2 & 32.8  \\
    \ \ w/o. preference & 39.8 & 50.8 & 53.6 & 52.2 & 12.5 & 22.5 & 23.3 & 22.9 & 65.7  & 27.2 & 30.4 & 28.7 \\
    \cdashlinelr{1-13}
    \rowcolor{gray!10}\ours (Iteration 3) & 44.2 & 76.2 & 84.2 & 80.0 & 9.6 & 62.4 & 69.1 & 65.6 & 69.6  & 60.0 & 56.6 & 58.2 \\
    \ \ w/o. warm-up & 28.6 & 67.3 & 58.2 & 62.4 & 6.4 & 46.8 & 38.4 & 42.2  & 70.4 & 44.9 & 39.2 & 41.9  \\
    \ \ w/o. preference & 40.7 & 55.7 & 58.3 & 57.0 & 11.9 & 25.3 & 26.2 & 25.7 & 67.8  & 31.3 & 33.5 & 32.4 \\
    \bottomrule
    \end{tabular}%
  }
  \caption{Ablation study results across three datasets over three iterations. We compare \ours with two variants: one that does not utilize synthetic data for initial warming-up (w/o warm-up) and another lacking fine-grained preference optimization for self-improvement (w/o preference).}
  \label{tab:iteration_result}
\end{table*}%

We provide the main results and the performance of \ours across different iterations in Table~\ref{tab:main_result}.
\paragraph{\ours effectively improves performance.} 
As shown in Table~\ref{tab:main_result}, \ours shows superior performance across three datasets and achieves \textit{state-of-the-art} results in citation quality. Specifically, \ours shows significant improvements over both ICL and Post-hoc approaches, highlighting the benefits of supervised signals in unlocking the attribution ability of LLMs.
Notably, compared with methods that rely on distilling from more advanced LLMs or training on human-annotated data, \ours achieves performance improvement of at least 8.0\%, 20.4\%, and 47.0\% in citation quality for ASQA, ELI5, and StrategyQA respectively. Regarding correctness, \ours also achieves gains of at least 9.1\% and 7.2\% on both ASQA and StrategyQA, despite a slight decrease on ELI5.

\paragraph{\ours successfully achieves self-improvement.}
We compare the performance of \ours from iteration 0 to 3 in Table~\ref{tab:main_result}, and the results demonstrate consistent improvements across iterations. Initially, at iteration 0 (after warm-up), thanks to the synthetic training data, the model shows decent performance after warm-up.
By iteration 1, \ours exhibits remarkable effectiveness in improving its performance by leveraging its own generated samples (\eg, $ 23.5\to72.0 $ on ASQA, $ 10.0\to48.9 $ on ELI5, $ 9.5\to46.4 $ on StrategyQA). Subsequent iterations continue this trend of incremental improvement, reaching a convergence point at iteration 3.
\subsection{Ablation Study and Analysis}
We conduct comprehensive ablation studies and analyses to understand how each component in \ours contributes to the significant improvement.
\paragraph{Effect of synthetic data warming-up.}
\begin{table}[!t]
\small
\centering
\resizebox{\linewidth}{!}{
\begin{tabular}{@{}lcccc@{}}
\toprule
\textbf{Model}  & \textbf{Iteration 1} & \textbf{Iteration 2} & \textbf{Iteration 3}\\ 
\midrule
\ours & 42.5\% & 90.2\% & 95.9\% \\
\ \ w/o. warm-up & 3.24\% & 41.2\% & 83.8\% \\
\bottomrule
\end{tabular}
}
\caption{The pass rate comparison between \ours and \ours (w/o. warm-up) across different iterations during the rejection sampling stage.}
\vspace{-4mm}
\label{tab:ablation_warm_up}
\end{table}

\definecolor{tiffanyblue}{RGB}{129,216,208}
\definecolor{bangdiblue}{RGB}{0,149,182}
\definecolor{kleinblue}{RGB}{0,47,167}
\definecolor{kabuliblue}{RGB}{26,85,153}
\definecolor{purple}{RGB}{138,43,226}

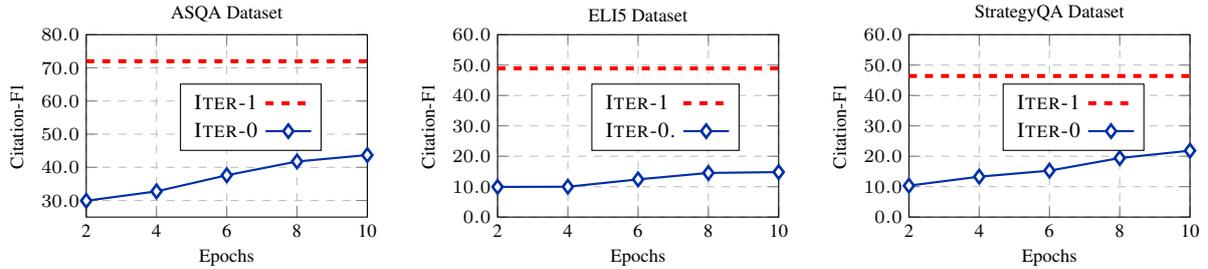
\begin{figure*}[htbp]
  \centering
  \begin{tikzpicture}[]
    \scriptsize{
      \begin{axis}[
      title={ASQA Dataset},
      title style={yshift=-0.5em},
	 at={(0em,13em)},
      ymajorgrids,
      xmajorgrids,
      grid style=dashed,
      width=.33\textwidth,
      height=.25\textwidth,
      legend style={at={(0.4,0.45)}, anchor=south west},
      xlabel={\scriptsize{Epochs}},
      ylabel={\scriptsize{Citation-F1}},
      ylabel style={yshift=0.25em},xlabel style={yshift=0.0em},
      yticklabel style={/pgf/number format/precision=1,/pgf/number format/fixed zerofill},
      ymin=25,ymax=80, ytick={20,30,40,50,60,70,80},
      xmin=2,xmax=10,xtick={2,4,6,8,10},
      xticklabels={2, 4, 6, 8, 10},
      legend style={yshift=-6pt,xshift=-1em, legend plot pos=right,font={\footnotesize},cells={anchor=west}}
      ]

      \addplot[red,thick,dashed, line width=1.5pt] coordinates {(2,72.0) (10, 72.0)};
      \addlegendentry{\scalebox{.9}{\textsc{Iter-1}}}

      \addplot[kleinblue,mark=diamond*,mark size=2.5pt,thick,mark options={fill=white,draw=kleinblue,line width=1.0pt}] coordinates {(2,29.92) (4,32.75) (6,37.61) (8,41.76) (10, 43.70)
      };
      \addlegendentry{\scalebox{.9}{\textsc{Iter-0}}}

      \end{axis}
     }
	\scriptsize{
      \begin{axis}[
      title={ELI5 Dataset},
      title style={yshift=-0.5em},
	 at={(22em,13em)},
      ymajorgrids,
      xmajorgrids,
      grid style=dashed,
      width=.33\textwidth,
      height=.25\textwidth,
      legend style={at={(0.4,0.45)}, anchor=south west},
      xlabel={\scriptsize{Epochs}},
      ylabel={\scriptsize{Citation-F1}},
      ylabel style={yshift=0.25em},xlabel style={yshift=0.0em},
      yticklabel style={/pgf/number format/precision=1,/pgf/number format/fixed zerofill},
      ymin=0,ymax=60, ytick={0,10,20,30,40,50,60},
      xmin=2,xmax=10,xtick={2,4,6,8,10},
      xticklabels={2, 4, 6, 8, 10},
      legend style={yshift=-6pt,xshift=-1em, legend plot pos=right,font={\footnotesize},cells={anchor=west}}
      ]
      \addplot[red,thick,dashed, line width=1.5pt] coordinates {(2,48.9) (10, 48.9)};
      \addlegendentry{\scalebox{.9}{\textsc{Iter-1}}}

      \addplot[kleinblue,mark=diamond*,mark size=2.5pt,thick,mark options={fill=white,draw=kleinblue,line width=1.0pt}] coordinates {(2,9.92)  (4,9.98) (6,12.40) (8,14.52) (10,14.79)};
      \addlegendentry{\scalebox{.9}{\textsc{Iter-0.}}}
      \end{axis}
     }
     
  	\scriptsize{
      \begin{axis}[
      title={StrategyQA Dataset},
      title style={yshift=-0.5em},
	 at={(44em,13em)},
      ymajorgrids,
      xmajorgrids,
      grid style=dashed,
      width=.33\textwidth,
      height=.25\textwidth,
      legend style={at={(0.4,0.45)}, anchor=south west},
      xlabel={\scriptsize{Epochs}},
      ylabel={\scriptsize{Citation-F1}},
      ylabel style={yshift=0.25em},xlabel style={yshift=0.0em},
      yticklabel style={/pgf/number format/precision=1,/pgf/number format/fixed zerofill},
      ymin=0,ymax=60, ytick={0,10,20,30,40,50,60},
      xmin=2,xmax=10,xtick={2,4,6,8,10},
      xticklabels={2, 4, 6, 8, 10},
      legend style={yshift=-6pt,xshift=-1em, legend plot pos=right,font={\footnotesize},cells={anchor=west}}
      ]
	\addplot[red,thick,dashed, line width=1.5pt] coordinates {(2,46.4) (10, 46.4)};
      \addlegendentry{\scalebox{.9}{\textsc{Iter-1}}}

      \addplot[kleinblue,mark=diamond*,mark size=2.5pt,thick,mark options={fill=white,draw=kleinblue,line width=1.0pt}] coordinates {(2,10.34) (4,13.29) (6,15.27) (8,19.42) (10,21.84)
      };
      \addlegendentry{\scalebox{.9}{\textsc{Iter-0}}}

      \end{axis}
     }
	\end{tikzpicture}
  \caption{The impact of supervision signals from different stages (synthetic data v.s. self-improvement) on attribution performance across ASQA, ELI5, and StrategyQA. The blue line represents the model that undergoes only supervised fine-tuning use synthetic data at iteration 0. The red line represents the model that first trains for two epochs with synthetic data at iteration 0, followed by one iteration of self-improvement.}
  \label{fig:more_epochs}
\end{figure*}
\definecolor{Maroon}{HTML}{AE3135}
\definecolor{BLUE}{HTML}{6466AE}
\definecolor{my-green}{HTML}{8ECFC9}
\definecolor{my-yellow}{HTML}{FFBE7A}
\definecolor{my-blue}{HTML}{82B0D2}

\begin{figure}[t]
    \centering
\pgfplotsset{width=0.6\linewidth,height=0.55\linewidth,compat=1.15}
\footnotesize
\begin{tikzpicture}
\scriptsize{
\begin{axis}[
	at={(0em,0em)},
    xlabel={Synthetic data sizes},
    ylabel={Citation F1 (\%)},
    xmin=0.05, xmax=0.55,
    ymin=43, ymax=87,
    xtick={0.1, 0.3, 0.5},
    ytick={45, 55, 65, 75, 85},
    ymajorgrids=true,
    xmajorgrids=true,
    grid style=dashed,
    xticklabels={1k, 3k, 5k},
    x label style={at={(axis description cs:0.5,-0.125)},anchor=north},
    y label style={at={(axis description cs:-0.15,0.5)},anchor=south},
    legend style={
    	at={(1.0,1.03)},
    	anchor=south,
    	legend columns=-1,
    	nodes={scale=0.7, transform shape}}
]
\addplot[
    color=Maroon,
    mark=diamond*,
    mark size=2.5pt,thick,line width=2.2pt,
    mark options={fill=Maroon,draw=Maroon,line width=2.2pt}
    ]
    coordinates {
    (0.1, 75.4)
    (0.3, 79.4)
    (0.5, 80.9)
    };
    \addlegendentry{ASQA}
\addplot[
    color=BLUE,
    mark=pentagon*,
    mark size=2.5pt,thick,line width=2.2pt,
    mark options={fill=BLUE,draw=BLUE,line width=2.2pt}
    ]
    coordinates {
    (0.1, 53.6)
    (0.3, 60.6)
    (0.5, 65.6)
    };
    \addlegendentry{ELI5}
\addplot[
    color=my-yellow,
    mark=*,
    mark size=2.5pt,thick,line width=2.2pt,
    mark options={fill=my-yellow,draw=my-yellow,line width=2.2pt}
    ]
    coordinates {
    (0.1, 50.8)
    (0.3, 55.6)
    (0.5, 58.2)
    };
    \addlegendentry{StrategyQA}
\end{axis}

\begin{axis}[
	at={(16em,0em)},
    xlabel={Synthetic data sizes},
    ylabel={Correctness (\%)},
    xmin=0.05, xmax=0.55,
    ymin=0, ymax=77,
    xtick={0.1, 0.3, 0.5},
    ytick={0, 15, 30, 45, 60, 75},
    ymajorgrids=true,
    xmajorgrids=true,
    grid style=dashed,
    xticklabels={1k, 3k, 5k},
    x label style={at={(axis description cs:0.5,-0.125)},anchor=north},
    y label style={at={(axis description cs:-0.15,0.5)},anchor=south}
]
\addplot[
    color=Maroon,
    mark=diamond*,
    mark size=2.5pt,thick,line width=2.2pt,
    mark options={fill=Maroon,draw=Maroon,line width=2.2pt}
    ]
    coordinates {
    (0.1, 43.3)
    (0.3, 42.0)
    (0.5, 43.1)
    };
\addplot[
    color=BLUE,
    mark=pentagon*,
    mark size=2.5pt,thick,line width=2.2pt,
    mark options={fill=BLUE,draw=BLUE,line width=2.2pt}
    ]
    coordinates {
    (0.1, 10.1)
    (0.3, 10.6)
    (0.5, 10.2)
    };
\addplot[
    color=my-yellow,
    mark=*,
    mark size=2.5pt,thick,line width=2.2pt,
    mark options={fill=my-yellow,draw=my-yellow,line width=2.2pt}
    ]
    coordinates {
    (0.1, 61.2)
    (0.3, 64.6)
    (0.5, 72.7)
    };
\end{axis}
}
\end{tikzpicture}
    \caption{Ablation study on the effect of synthetic data size on attribution and correctness performance. We sample 1k, 3k, and 5k user queries for data synthesis.
    }
    \vspace{-8mm}
    \label{fig:training_size}
\end{figure}
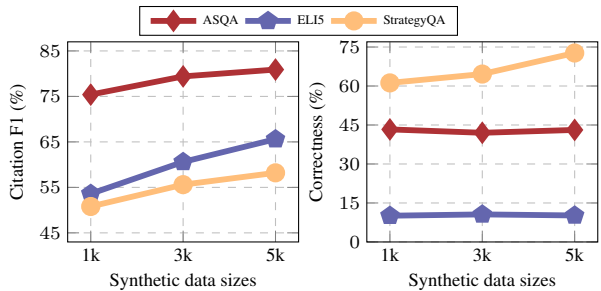
To demonstrate the importance of utilizing synthetic data for initial warm-up in \ours, we conduct a comparative ablation study employing \texttt{Llama-2-13b} for self-improvement, omitting the initial warm-up stage.
Table~\ref{tab:iteration_result} shows the ablation results (w/o. warm-up) across three iterations. 
We observe that omitting the initial warm-up stage can lead to a significant performance drop in the first iteration.
Additionally, as the iteration increases, the performance of the model without warm-up shows only modest improvements and remains substantially inferior to the model that underwent warm-up. 
Moreover, we also calculate the pass rate of sampled response in each iteration as shown in Table~\ref{tab:ablation_warm_up}. 
The findings indicate that the model with warm-up exhibits a higher pass rate in the first iteration, which allows the model to utilize more supervised signals for self-improvement.
These results suggest that warming up effectively facilitates the bootstrapping of supervised data, thus preventing early model stagnation.
It's worth noting that while the warm-up strategy effectively enriches the model with supervision signals at an early stage, it does not lead to noticeable improvements in citation quality, as shown in Table~\ref{tab:main_result}. We hypothesize that this limitation stems from the inherent difficulty LLMs face in synthesizing information from multiple sources to generate comprehensive and attributable responses solely through direct supervised fine-tuning.
\paragraph{Effect of fine-grained preference optimization.}
To further understand the significance of fine-grained preference optimization, we compare an ablation of \ours that solely relies on high-quality samples for iteratively supervised fine-tuning, discarding low-quality samples for fine-grained preference optimization.
As shown in Table~\ref{tab:iteration_result}, there is a significant decline in performance when fine-grained preference optimization is removed. This highlights the effectiveness of \ours in fully unlocking the potential of low-quality samples to enhance attribution performance.

\paragraph{Effect of synthetic data size.}
We investigate the effect of varying synthetic data sizes on the performance of \ours. Figure~\ref{fig:training_size} demonstrates their effect on citation quality and correctness after three iterations of self-improving. 
Specifically, we sample 1k, 3k, and 5k unlabeled queries to generate synthetic training data accordingly, which provides different levels of supervision signals.
As shown in Figure~\ref{fig:training_size}, even with 1k synthetic data points, \ours demonstrates comparable performance.
Moreover, as the training size increases, \ours achieves notable improvement in citation quality and exhibits stability in correctness.
\paragraph{Supervision signals from synthetic data v.s. iterative self-improvement.}
We further investigate the differential impact of supervision signals derived from data synthesis versus those from the iterative self-improvement stage.
We utilize synthetic training data to train the model for multiple epochs, extending up to 10 epochs, and compare its performance to that of a model that undergoes only the first iteration of self-improvement.
As depicted in Figure~\ref{fig:more_epochs}, training with synthetic data during the initial iteration yields minimal performance gains. The attribution performance climbs slowly as training epochs increase and fails to surpass the performance of the model after just one iteration of self-improvement. This observation reveals the importance of the supervision signals provided by the model itself during self-improvement.
\section{Human Evaluation}
\label{sec:analysis}
\begin{table}[t]
\small
\centering
\resizebox{\linewidth}{!}{
\begin{tabular}{@{}lccccc@{}}
\toprule
& \multicolumn{3}{c}{Attribution} & \multicolumn{2}{c}{Overall Quality} \\
\cmidrule(lr){2-4} \cmidrule(lr){5-6}
& Full & Partial & No & Corr. & Comp.\\
\midrule
ChatGPT (\textsc{ICL})   & \ul{68.5\%} & \ul{22.1\%} & \ul{9.4\%}   & \textbf{3.6} & \ul{4.4}   \\
Distill-Llama-3-70B-Instruct & 54.6\% & 32.4\% & 13.0\% & 2.9 & 3.2 \\
Self-RAG \citep{asai2023self} & 45.7\% & 27.5\% & 26.8\% & 2.4 & 2.1 \\
FGR \citep{huang2024training} & 58.4\% & 28.7\% & 12.9\% & 2.5 & 2.8 \\
\ours (Ours)    & \textbf{76.2\%}  & \textbf{18.3\%} & \textbf{5.5\%} & \ul{3.5} & \textbf{4.6}   \\
\bottomrule
\end{tabular}
}
\caption{Human evaluation results on attribution, correctness (\textbf{Corr.}), and comprehensiveness (\textbf{Comp.}). \textbf{Bold} numbers indicate the best performance, while ``$\_$'' indicates the second-best performance.}
\vspace{-4mm}
\label{tab:human_evaluation}
\end{table}

Human evaluation results, detailed in Table~\ref{tab:human_evaluation}, indicate that \ours generates significantly more attributable responses compared to all baselines, even surpassing ChatGPT\footnote{We utilize \texttt{gpt-3.5-turbo-0125} version.}.
Specifically, 76.2\% of the statements generated by \ours are fully supported by the cited documents, which outperforms ChatGPT by 11.24\%.
Additionally, 18.3\% of the statements are partially supported, with only 5.5\% unsupported.
In terms of factuality, \ours outperforms all training-based baselines, slightly inferior to ChatGPT.
Moreover, \ours achieves the highest score in comprehensiveness, demonstrating its exceptional ability to generate responses that extensively cover information from multiple sources.
Overall, these findings are in line with the automatic evaluation results in Table~\ref{tab:main_result}.

\section{Conclusion}
We propose \ours, a self-improvement framework to push the frontier of LLM attribution. We identify two key limitations for LLM attribution self-improvement. To address these, \ours first leverages self-constructed synthetic data for warming up, aiming to prevent models from early stagnation due to insufficient supervision signals.  
To explore more fine-grained supervision signals, \ours constructs fine-grained preference supervision signals from low-quality samples for preference optimization.
Both automatic and human evaluations demonstrate significant improvement in attribution without relying on human annotations and more advanced LLMs. 
\section*{Limitations}
\label{sec:limitations}
Despite significant performance improvements, our work presents several limitations worth noting.
\textbf{Firstly}, while our data synthesis process provides a good starting point for the model to self-improve and demonstrate some generalization on existing benchmarks, it may not cover all scenarios encountered in user information-seeking. This limitation raises concerns regarding the generalizability of synthetic data in a more complex information-seeking environment.
\textbf{Secondly}, the iterative training pipeline of our self-improving framework is time-consuming, presenting a significant trade-off between performance and training duration.
\textbf{Thirdly}, although our self-improving framework does not rely on human annotations and more advanced LLMs, it still necessitates the integration of off-the-shelf NLI models to guarantee the quality of attribution in the generated samples. The performance of the NLI model significantly impacts the quality of our outputs to a certain extent. To move towards a fully self-improving framework that does not rely on external judgment, future research could investigate the use of intrinsic attribution signals derived directly from the LLM itself.

\section*{Acknowledgements}
Xiaocheng Feng is the corresponding author of this work. We thank the anonymous reviewers for their insightful comments. This work was supported by the National Natural Science Foundation of China (NSFC) (grant 62276078, U22B2059), the Key R\&D Program of Heilongjiang via grant 2022ZX01A32, the International Cooperation Project of PCL, PCL2022D01 and the Fundamental Research Funds for the Central Universities (Grant No.HIT.OCEF.2023018).

\bibliography{custom}

\appendix
\definecolor{WindowsBlue}{RGB}{2,152,219}
\section{Data Synthesis}
\label{appendix:data_synthesis}
\subsection{Data Sources}
The queries employed for data synthesis are sourced from the Wish-QA \citep{yehudai2024genie}, which provides high-quality grounded data suitable for content-grounded generation tasks such as long-form question-answering and summarization.
Specifically, we utilize the ELI5 subset of the WishQA, noted for its high lexical diversity, comprising a total of 8,413 queries. Notably, we randomly sample 5,000 user queries for our data synthesis, resulting in the creation of 5,000 synthetic data points.

\subsection{Prompts for Data Synthesis}
\begin{figure*}
    \centering
    \footnotesize{
\begin{tikzpicture}

\begin{scope}[]

\node [anchor=south west] (n1) at (0, 0) {};

\node [anchor=north,rectangle,rounded corners=10pt,minimum height=1.1in,minimum width=6.0in, draw, line width=1.5pt, draw=black!60!white] (b1) at ([xshift=0em, yshift=-1em]n1.south) {};

\node [anchor=center,rectangle,rounded corners=5pt,minimum height=1em,minimum width=2em,fill=teal!30, draw=black!60!white, line width=1.3pt] (l1) at ([xshift=0em, yshift=0em]b1.north) {\small{(a) Prompt template for response generation}};

\node [anchor=north west,rectangle,rounded corners=2pt,minimum height=1em,minimum width=2em, text width=5.8in] (n21) at ([xshift=0.5em, yshift=-1em]b1.north west) {\texttt{\textbf{Instruction:} Given a question, generate a detailed and informative response that covers multiple perspectives and synthesizes information from various sources. Limit the response to a maximum of five statements.}};
\node [anchor=north west,rectangle,rounded corners=2pt,minimum height=1em,minimum width=2em, text width=5.8in] (n22) at ([xshift=0em, yshift=-0.2em]n21.south west) {\texttt{\textbf{Question:} [Question]}};
\node [anchor=north west,rectangle,rounded corners=2pt,minimum height=1em,minimum width=2em, text width=5.8in] (n23) at ([xshift=0em, yshift=-0.2em]n22.south west) {\texttt{\textbf{Response:}}};

\begin{pgfonlayer}{background}
\node [anchor=north,rectangle,rounded corners=3pt,minimum height=0.9in,minimum width=5.85in,fill=gray!10] (bb) at ([xshift=0em, yshift=-0.2em]l1.south) {};

\end{pgfonlayer}

\end{scope}

\begin{scope}[yshift=-1.3in]

\node [anchor=south west] (n1) at (0, 0) {};

\node [anchor=north,rectangle,rounded corners=10pt,minimum height=1.1in,minimum width=6.0in, draw, line width=1.5pt, draw=black!60!white] (b1) at ([xshift=0em, yshift=-1em]n1.south) {};

\node [anchor=center,rectangle,rounded corners=5pt,minimum height=1em,minimum width=2em,fill=teal!30, draw=black!60!white, line width=1.3pt] (l1) at ([xshift=0em, yshift=0em]b1.north) {\small{(b) Prompt template for claim decomposition}};

\node [anchor=north west,rectangle,rounded corners=2pt,minimum height=1em,minimum width=2em, text width=5.8in] (n21) at ([xshift=0.5em, yshift=-1em]b1.north west) {\texttt{\textbf{Instruction:} Given a detailed and informative response, break it into its constituent claims. Identify and list each distinct claim, ensuring to capture all essential elements and nuances presented in the original response.}};
\node [anchor=north west,rectangle,rounded corners=2pt,minimum height=1em,minimum width=2em, text width=5.8in] (n22) at ([xshift=0em, yshift=-0.2em]n21.south west) {\texttt{\textbf{Response:} [Response]}};
\node [anchor=north west,rectangle,rounded corners=2pt,minimum height=1em,minimum width=2em, text width=5.8in] (n23) at ([xshift=0em, yshift=-0.2em]n22.south west) {\texttt{\textbf{Claims:}}};

\begin{pgfonlayer}{background}
\node [anchor=north,rectangle,rounded corners=3pt,minimum height=0.9in,minimum width=5.85in,fill=gray!10] (bb) at ([xshift=0em, yshift=-0.2em]l1.south) {};

\end{pgfonlayer}

\end{scope}

\begin{scope}[yshift=-2.6in]

\node [anchor=south west] (n1) at (0, 0) {};

\node [anchor=north,rectangle,rounded corners=10pt,minimum height=1.0in,minimum width=6.0in, draw, line width=1.5pt, draw=black!60!white] (b1) at ([xshift=0em, yshift=-1em]n1.south) {};

\node [anchor=center,rectangle,rounded corners=5pt,minimum height=1em,minimum width=2em,fill=teal!30, draw=black!60!white, line width=1.3pt] (l1) at ([xshift=0em, yshift=0em]b1.north) {\small{(c) Prompt template for document generation}};

\node [anchor=north west,rectangle,rounded corners=2pt,minimum height=1em,minimum width=2em, text width=5.8in] (n21) at ([xshift=0.5em, yshift=-1em]b1.north west) {\texttt{\textbf{Instruction:} Given a claim, generate a 100-word document with a title. The main content of the document should elaborate on the claims and contain the main content of the claim.}};
\node [anchor=north west,rectangle,rounded corners=2pt,minimum height=1em,minimum width=2em, text width=5.8in] (n22) at ([xshift=0em, yshift=-0.2em]n21.south west) {\texttt{\textbf{Claim:} [Claim]}};
\node [anchor=north west,rectangle,rounded corners=2pt,minimum height=1em,minimum width=2em, text width=5.8in] (n23) at ([xshift=0em, yshift=-0.2em]n22.south west) {\texttt{\textbf{Documents:}}};

\begin{pgfonlayer}{background}
\node [anchor=north,rectangle,rounded corners=3pt,minimum height=0.8in,minimum width=5.85in,fill=gray!10] (bb) at ([xshift=0em, yshift=-0.2em]l1.south) {};

\end{pgfonlayer}

\end{scope}
\end{tikzpicture}
}
\caption{Illustration of the prompting design for the data synthesis pipeline.}
\label{fig:data_synthesis_prompts}
\end{figure*}
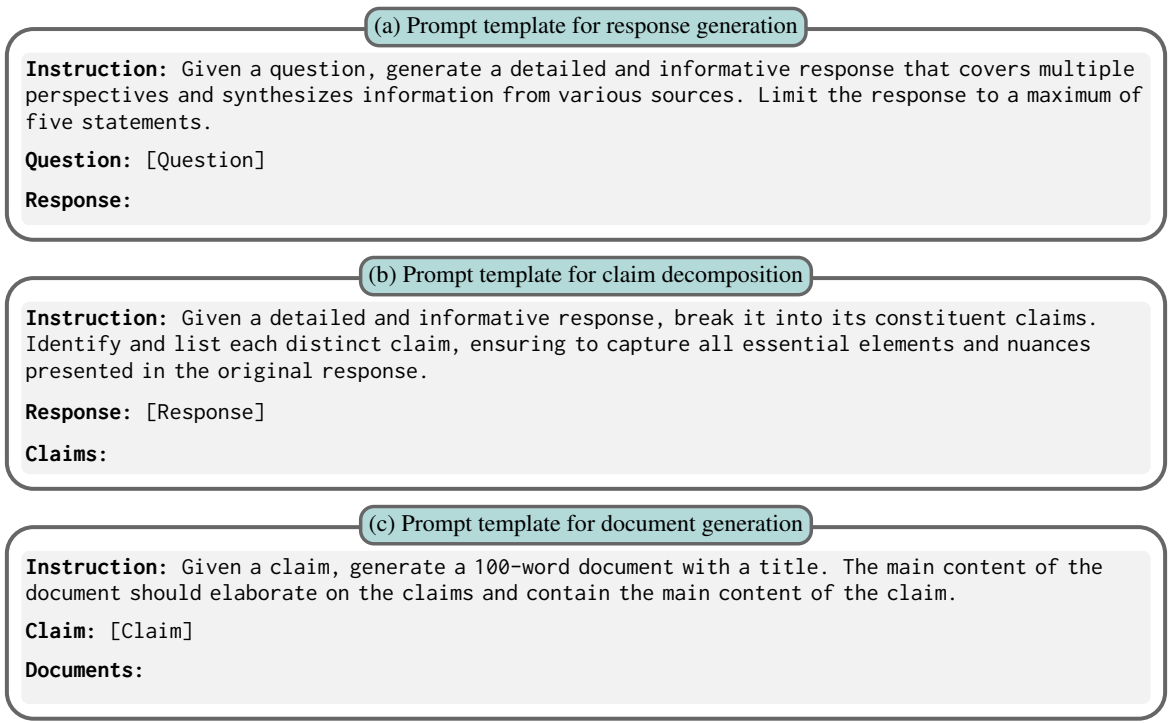

We detail the prompts employed in the synthetic data generation stage, covering response generation, claim decomposition, and document generation, shown in Figure~\ref{fig:data_synthesis_prompts}.

\subsection{Implementation Details}
In our work, we use \texttt{Llama-2-13b-base} for data synthesis, as our goal is to realize self-improving for the attribution ability of LLMs, the models used in the data synthesis stage and the subsequent main experiment need to be consistent without introducing additional more powerful models.
To enhance the LLM's ability to accurately follow instructions at each step, we utilize in-context learning, incorporating two demonstrations for response generation, claim decomposition, and document generation.

\subsection{Quality of Synthetic Data}
We focus on evaluating the attributability of the final response. Specifically, we employ an off-the-shelf Natural Language Inference (NLI) model, TRUE \citep{honovich2022true}, to verify whether each statement in the response is fully supported by the cited documents and to check for the presence of any irrelevant citations. The results indicate that the synthetic data are of significantly high quality: 92.3\% of the statements are fully supported by the cited documents, and 94.1\% are free from irrelevant citations.

\section{Details of evaluation datasets}
\label{appendix:datasets}
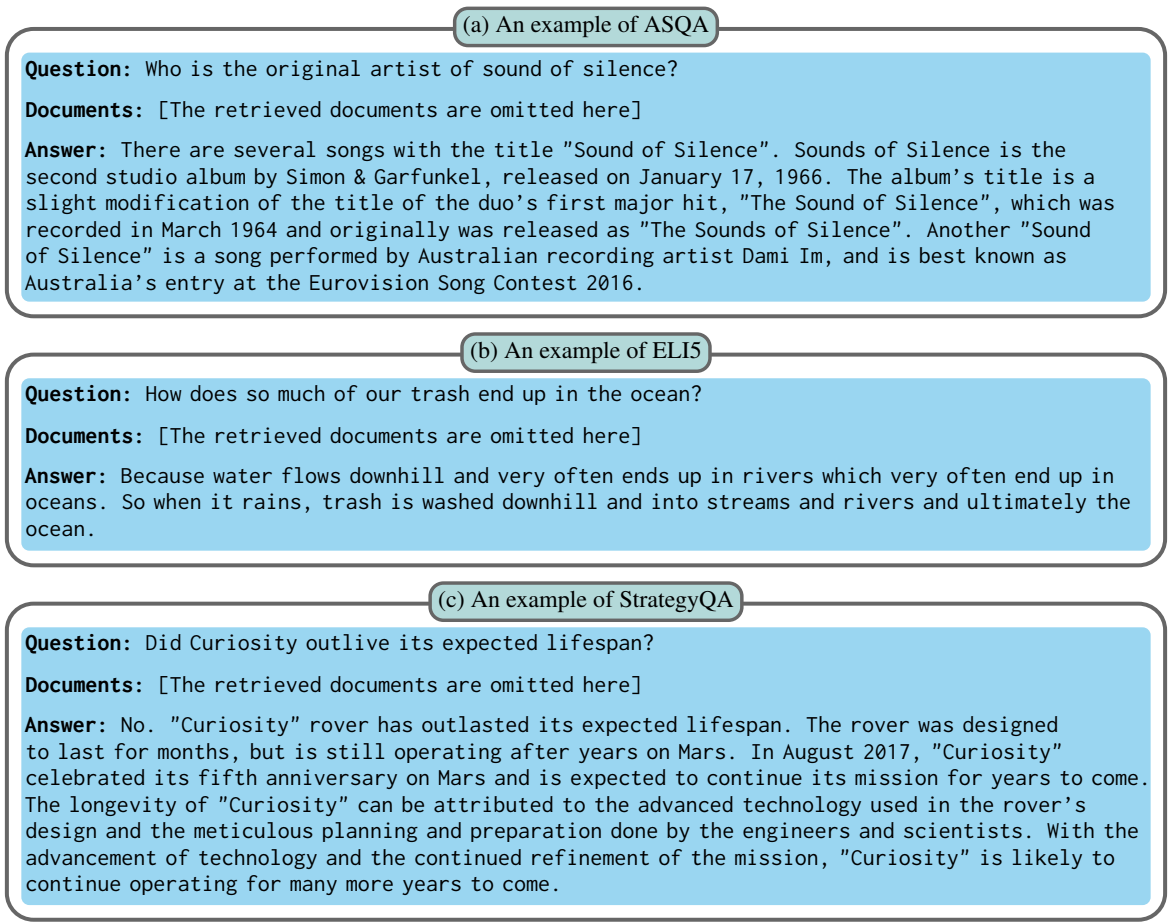
\begin{figure*}
    \centering
    \footnotesize{
\begin{tikzpicture}

\begin{scope}[]

\node [anchor=south west] (n1) at (0, 0) {};

\node [anchor=north,rectangle,rounded corners=10pt,minimum height=1.5in,minimum width=6.0in, draw, line width=1.5pt, draw=black!60!white] (b1) at ([xshift=0em, yshift=-1em]n1.south) {};

\node [anchor=center,rectangle,rounded corners=5pt,minimum height=1em,minimum width=2em,fill=teal!30, draw=black!60!white, line width=1.3pt] (l1) at ([xshift=0em, yshift=0em]b1.north) {\small{(a) An example of ASQA}};

\node [anchor=north west,rectangle,rounded corners=2pt,minimum height=1em,minimum width=2em, text width=5.8in] (n21) at ([xshift=0.5em, yshift=-1em]b1.north west) {\texttt{\textbf{Question:} Who is the original artist of sound of silence?}};
\node [anchor=north west,rectangle,rounded corners=2pt,minimum height=1em,minimum width=2em, text width=5.8in] (n22) at ([xshift=0em, yshift=-0.2em]n21.south west) {\texttt{\textbf{Documents:} [The retrieved documents are omitted here]}};
\node [anchor=north west,rectangle,rounded corners=2pt,minimum height=1em,minimum width=2em, text width=5.8in] (n23) at ([xshift=0em, yshift=-0.2em]n22.south west) {\texttt{\textbf{Answer:} There are several songs with the title "Sound of Silence". Sounds of Silence is the second studio album by Simon \& Garfunkel, released on January 17, 1966. The album's title is a slight modification of the title of the duo's first major hit, "The Sound of Silence", which was recorded in March 1964 and originally was released as "The Sounds of Silence". Another "Sound of Silence" is a song performed by Australian recording artist Dami Im, and is best known as Australia's entry at the Eurovision Song Contest 2016.}};

\begin{pgfonlayer}{background}
\node [anchor=north,rectangle,rounded corners=3pt,minimum height=1.3in,minimum width=5.85in,fill=WindowsBlue!40] (bb) at ([xshift=0em, yshift=-0.2em]l1.south) {};

\end{pgfonlayer}

\end{scope}

\begin{scope}[yshift=-1.7in]

\node [anchor=south west] (n1) at (0, 0) {};

\node [anchor=north,rectangle,rounded corners=10pt,minimum height=1.1in,minimum width=6.0in, draw, line width=1.5pt, draw=black!60!white] (b1) at ([xshift=0em, yshift=-1em]n1.south) {};

\node [anchor=center,rectangle,rounded corners=5pt,minimum height=1em,minimum width=2em,fill=teal!30, draw=black!60!white, line width=1.3pt] (l1) at ([xshift=0em, yshift=0em]b1.north) {\small{(b) An example of ELI5}};

\node [anchor=north west,rectangle,rounded corners=2pt,minimum height=1em,minimum width=2em, text width=5.8in] (n21) at ([xshift=0.5em, yshift=-1em]b1.north west) {\texttt{\textbf{Question:} How does so much of our trash end up in the ocean?}};
\node [anchor=north west,rectangle,rounded corners=2pt,minimum height=1em,minimum width=2em, text width=5.8in] (n22) at ([xshift=0em, yshift=-0.2em]n21.south west) {\texttt{\textbf{Documents:} [The retrieved documents are omitted here]}};
\node [anchor=north west,rectangle,rounded corners=2pt,minimum height=1em,minimum width=2em, text width=5.8in] (n23) at ([xshift=0em, yshift=-0.2em]n22.south west) {\texttt{\textbf{Answer:} Because water flows downhill and very often ends up in rivers which very often end up in oceans. So when it rains, trash is washed downhill and into streams and rivers and ultimately the ocean.}};

\begin{pgfonlayer}{background}
\node [anchor=north,rectangle,rounded corners=3pt,minimum height=0.9in,minimum width=5.85in,fill=WindowsBlue!40] (bb) at ([xshift=0em, yshift=-0.2em]l1.south) {};

\end{pgfonlayer}

\end{scope}

\begin{scope}[yshift=-3.0in]

\node [anchor=south west] (n1) at (0, 0) {};

\node [anchor=north,rectangle,rounded corners=10pt,minimum height=1.65in,minimum width=6.0in, draw, line width=1.5pt, draw=black!60!white] (b1) at ([xshift=0em, yshift=-1em]n1.south) {};

\node [anchor=center,rectangle,rounded corners=5pt,minimum height=1em,minimum width=2em,fill=teal!30, draw=black!60!white, line width=1.3pt] (l1) at ([xshift=0em, yshift=0em]b1.north) {\small{(c) An example of StrategyQA}};

\node [anchor=north west,rectangle,rounded corners=2pt,minimum height=1em,minimum width=2em, text width=5.8in] (n21) at ([xshift=0.5em, yshift=-1em]b1.north west) {\texttt{\textbf{Question:} Did Curiosity outlive its expected lifespan?}};
\node [anchor=north west,rectangle,rounded corners=2pt,minimum height=1em,minimum width=2em, text width=5.8in] (n22) at ([xshift=0em, yshift=-0.2em]n21.south west) {\texttt{\textbf{Documents:} [The retrieved documents are omitted here]}};
\node [anchor=north west,rectangle,rounded corners=2pt,minimum height=1em,minimum width=2em, text width=5.8in] (n23) at ([xshift=0em, yshift=-0.2em]n22.south west) {\texttt{\textbf{Answer:} No. "Curiosity" rover has outlasted its expected lifespan. The rover was designed to last for months, but is still operating after years on Mars. In August 2017, "Curiosity" celebrated its fifth anniversary on Mars and is expected to continue its mission for years to come. The longevity of "Curiosity" can be attributed to the advanced technology used in the rover's design and the meticulous planning and preparation done by the engineers and scientists. With the advancement of technology and the continued refinement of the mission, "Curiosity" is likely to continue operating for many more years to come.}};

\begin{pgfonlayer}{background}
\node [anchor=north,rectangle,rounded corners=3pt,minimum height=1.45in,minimum width=5.85in,fill=WindowsBlue!40] (bb) at ([xshift=0em, yshift=-0.2em]l1.south) {};

\end{pgfonlayer}

\end{scope}
\end{tikzpicture}
}
\caption{Examples of the ASQA, ELI5, and StrategyQA datasets.}
\label{fig:dataset_example}
\end{figure*}

Our evaluation utilizes the ASQA, ELI5, and StrategyQA datasets. 
For both ASQA and StrategyQA, Wikipedia serves as the external knowledge base, specifically employing the Wikipedia snapshot from 2018-12-20. For the ELI5 dataset, the external knowledge source is Sphere \citep{piktus2021the}. Regarding the retrievers, we use the dense retriever GTR \citep{ni2022large} for Wikipedia and the sparse retriever BM25 for Sphere.
Detailed statistics for these datasets are presented in Table~\ref{tab:dataset}.
In line with previous research by \citet{gao2023enabling}, we use the same evaluation datasets for ASQA and ELI5.
Regarding StrategyQA, we adopt the settings of \citet{ye2023effective}, utilizing a randomly split subset of 490 test instances for evaluation.
To further clarify, we provide an example from each dataset in Table~\ref{fig:dataset_example}.
\begin{figure*}
    \centering
    \footnotesize{
\begin{tikzpicture}

\begin{scope}[]

\node [anchor=south west] (n1) at (0, 0) {};

\node [anchor=north,rectangle,rounded corners=10pt,minimum height=1.5in,minimum width=6.0in, draw, line width=1.5pt, draw=black!60!white] (b1) at ([xshift=0em, yshift=-1em]n1.south) {};

\node [anchor=center,rectangle,rounded corners=5pt,minimum height=1em,minimum width=2em,fill=teal!30, draw=black!60!white, line width=1.3pt] (l1) at ([xshift=0em, yshift=0em]b1.north) {\small{(a) Prompt template of ASQA and ELI5}};

\node [anchor=north west,rectangle,rounded corners=2pt,minimum height=1em,minimum width=2em, text width=5.8in] (n21) at ([xshift=0.5em, yshift=-1em]b1.north west) {\texttt{\textbf{Instruction:} Write an accurate, engaging, and concise answer for the given question using only the provided search results (some of which might be irrelevant) and cite them properly. Use an unbiased and journalistic tone. Always cite for any factual claim. When citing several search results, use [1][2][3]. Cite at least one document and at most three documents in each sentence. If multiple documents support the sentence, only cite a minimum sufficient subset of the documents.}};
\node [anchor=north west,rectangle,rounded corners=2pt,minimum height=1em,minimum width=2em, text width=5.8in] (n22) at ([xshift=0em, yshift=-0.2em]n21.south west) {\texttt{\textbf{Question:} [Question]}};
\node [anchor=north west,rectangle,rounded corners=2pt,minimum height=1em,minimum width=2em, text width=5.8in] (n23) at ([xshift=0em, yshift=-0.2em]n22.south west) {\texttt{\textbf{Documents:} [Documents]}};

\begin{pgfonlayer}{background}
\node [anchor=north,rectangle,rounded corners=3pt,minimum height=1.3in,minimum width=5.85in,fill=gray!10] (bb) at ([xshift=0em, yshift=-0.2em]l1.south) {};

\end{pgfonlayer}

\end{scope}

\begin{scope}[yshift=-1.7in]

\node [anchor=south west] (n1) at (0, 0) {};

\node [anchor=north,rectangle,rounded corners=10pt,minimum height=1.1in,minimum width=6.0in, draw, line width=1.5pt, draw=black!60!white] (b1) at ([xshift=0em, yshift=-1em]n1.south) {};

\node [anchor=center,rectangle,rounded corners=5pt,minimum height=1em,minimum width=2em,fill=teal!30, draw=black!60!white, line width=1.3pt] (l1) at ([xshift=0em, yshift=0em]b1.north) {\small{(c) Prompt template of StrategyQA}};

\node [anchor=north west,rectangle,rounded corners=2pt,minimum height=1em,minimum width=2em, text width=5.8in] (n21) at ([xshift=0.5em, yshift=-1em]b1.north west) {\texttt{\textbf{Instruction:} Answer ``yes'' or ``no'' first. Then, write a clear and concise answer that combines reasoning with relevant search results and cite the sources properly, even if some might be irrelevant.}};
\node [anchor=north west,rectangle,rounded corners=2pt,minimum height=1em,minimum width=2em, text width=5.8in] (n22) at ([xshift=0em, yshift=-0.2em]n21.south west) {\texttt{\textbf{Question:} [Question]}};
\node [anchor=north west,rectangle,rounded corners=2pt,minimum height=1em,minimum width=2em, text width=5.8in] (n23) at ([xshift=0em, yshift=-0.2em]n22.south west) {\texttt{\textbf{Documents:} [Documents]}};

\begin{pgfonlayer}{background}
\node [anchor=north,rectangle,rounded corners=3pt,minimum height=0.9in,minimum width=5.85in,fill=gray!10] (bb) at ([xshift=0em, yshift=-0.2em]l1.south) {};

\end{pgfonlayer}

\end{scope}
\end{tikzpicture}
}
\caption{Illustration of the prompting design of evaluation datasets.}
\label{fig:prompt_ablations}
\end{figure*}
\begin{table}[htbp]
\small
\centering
\begin{tabular}{@{}lccc@{}}
\toprule
\textbf{Dataset}  & \textbf{Source} & \textbf{\# Examples}\\ 
\midrule
ASQA \citep{stelmakh2022asqa}   & Wiki   & 948    \\
ELI5 \citep{fan2019eli5}   & Sphere   & 1000    \\
StrategyQA \citep{geva2021did}    & Wiki  & 490    \\
\bottomrule
\end{tabular}
\caption{Statistics of datasets used for evaluation.}
\vspace{-4mm}
\label{tab:dataset}
\end{table}

\section{Automatic Evaluation Details}
\label{appendix:automatic_evaluation}

We provide a detailed description of the evaluation metrics employed to assess the quality of the model-generated responses.
\paragraph{Citation Quality.}
Citation Quality is a critical evaluation dimension in attributed text generation, assessing whether the answer is fully supported by the cited documents and that no irrelevant documents are cited. Following \citet{liu2023evaluating} and \citet{gao2023enabling}, the evaluation of citation quality is typically divided into two parts: \textbf{Citation Recall} and \textbf{Citation Precision}.

Citation Recall evaluates whether all generated statements are fully supported by the cited documents. Specifically, for each statement $s_i \in \mathcal{S}$, its citation recall is scored as 1 if there is at least one valid citation ($\mathcal{C}_i\neq \emptyset$) and the concatenation of cited documents $\text{concat}(\mathcal{C}_i)$ fully support the statement ($\phi(\text{concat}(\mathcal{C}_i), s_i)$ = 1), where $\phi(\text{premise}, \text{hypothesis})$ is an NLI model that outputs 1 if the premise entails the hypothesis. The final citation recall is calculated by averaging over all statements in $\mathcal{S}$.

Citation Precision assesses whether any citations in the response are irrelevant. A citation $c_{i,j}$ is determined as ``irrelevant'' if (a) $c_{i,j}$ alone cannot support statement $s_i$ and (b) removing $c_{i,j}$ does not affect the rest of the citations to support $s_i$.

Citation F1 is a metric that combines citation precision and citation recall by calculating their harmonic mean. In our work, we utilize this metric to evaluate the overall citation quality of the response, where a higher \textit{Citation F1} score indicates a more accurately and comprehensively attributed response.

\begin{equation}
	\begin{aligned}
    F_1 = 2 \cdot \frac{\textrm{citation precision} \cdot \textrm{citation recall}}{\textrm{citation precision} + \textrm{citation recall}},
    \label{eqn:citation_f1}
    \end{aligned}
\end{equation}

\paragraph{Correctness.}
Correctness is crucial in long-form QA tasks. Given the ambiguous nature of the ASQA dataset, where each question requires multiple short answers to cover different aspects, we follow \citet{stelmakh2022asqa} and calculate the recall of correct short answers using exact match.

As for the ELI5 dataset, evaluating the correctness of long-form answers is challenging. Thus, the ALCE benchmark employs InstructGPT (\texttt{text-davinci-003}) to generate three "sub-claims" based on the human-annotated answers. To assess correctness, we use a T5-11B model\footnote{\url{https://huggingface.co/google/t5_xxl_true_nli_mixture}} that has been fine-tuned on a collection of NLI datasets to check whether the model-generated outputs entail these sub-claims.

\section{Human Evaluation Details}
\label{appendix:human_evaluation}
Considering the open-ended nature of long-form QA tasks, automatic evaluation of correctness may not cover all possible answers. Furthermore, the evaluation of citation quality is constrained by the capabilities of the off-the-shelf NLI model, which may not adequately detect cases of \textit{partial support}. Therefore, we conduct a human evaluation to assess the attribution quality and correctness of \ours. We recruited two annotators, holding at least a bachelor's degree to participate in our study.

To evaluate citation quality, annotators are asked to verify whether each statement in the responses is fully supported, partially supported, or not supported by the cited documents and identify error types if the statement is not fully supported.

Next, we evaluate the overall quality of the responses, focusing on comprehensiveness and correctness. Annotators are asked to rate both comprehensiveness and correctness using a 5-point Likert scale, capturing different levels of content coverage and factuality.

\section{Baselines}
\label{appendix:baselines}
\paragraph{Knowledge Distillation:} We employ supervised fine-tuning to teach Llama-2-13B to generate responses with citations, utilizing training data distilled from the most advanced LLMs. Specifically, the queries and documents are sourced from our synthetic dataset and the attributed responses are generated by Llama-3-70B-Instruct / Mixtral-8x7B-Instruct.
\paragraph{Self-RAG \citep{asai2023self}:} The method involves training the LLM to generate text with reflection tokens, which are categorized into retrieval and critique tokens to indicate the need for retrieval and the attributability of its generation, respectively. Specifically, it first collects over 145,619 supervised data by prompting GPT-4 with specific instructions to generate responses with reflection tokens for knowledge-intensive queries. These data are then used to train the LLM to generate responses with self-reflection via supervised fine-tuning.
\paragraph{AGREE \citep{ye2023effective}:} The method involves training the LLM to generate grounded claims with citations and to identify unverified claims.  Specifically, it first collects 4,500 attribution data via post-hoc attribution with the help of an NLI model. These data are then used to train the model to generate grounded responses with citations and also clearly state the unsupported statements. An iterative retrieval process is employed to search for additional information for the unsupported statements via a test-time adaptation (TTA) strategy.
\paragraph{APO \citep{li2024improving}:} This method models the attributed text generation task as a preference learning task. Specifically, the model is first trained using 6,330 human-labeled high-quality attribution data for supervised fine-tuning to learn the basic ability of attribution. It then leverages automatically constructed preference data for preference learning, where a positive response is generated from relevant documents accompanied by a positive prompt, while a negative response is generated using irrelevant documents or a negative prompt.
\paragraph{FGR \citep{huang2024training}:} The method first collects 3,000 in-domain user queries along with retrieved documents and then leverages ChatGPT to generate high-quality attributed responses. These data then serve as training data to teach the model the basic ability of citation generation via supervised fine-tuning. Subsequently, the method designs reward models to teach the model to generate well-supported and accurate responses via fine-grained reinforcement learning.

To ensure a fair comparison, we employ the same base model (\texttt{Llama-2-13b-base}) for evaluating all baselines. For Self-RAG, AGREE, and APO, we directly utilize their published experimental results. In the case of FGR, which does not provide \texttt{Llama-2-13b-base} results, we reproduce the experiments using the official code and the same settings provided by the authors.

\section{Implement Details}
\label{appendix:implement}
In all experiments, training is conducted using eight A100-80GB GPUs, leveraging Deepspeed stage 3 \citep{rasley2020deepspeed} for multi-GPU distributed training, with training precision Bfloat16 enabled. 

During the initial warm-up stage, we employ the AdamW \citep{loshchilov2019decoupled} optimizer with a warm-up ratio of 0.03. The total batch size is set at 64, and the learning rate is maintained at 2e-5. The maximum input sequence length is configured to 2048 tokens. The model is trained with only 20\% of the synthetic dataset for two epochs in this stage. This strategy is designed to prevent the model from overfitting to the synthetic data during the warm-up stage, enabling it to generate more diverse samples in the subsequent rejection sampling fine-tuning stage.
In the self-improving stage, we conduct rejection-sampling fine-tuning for three epochs at each iteration, maintaining the same training settings as those used during the warming-up stage. To get the highest quality responses during rejection sampling, we set the threshold for attributability reward at 1.0, ensuring that every statement in the response is fully supported by the cited documents. For comprehensive, we set the threshold to 0.8, which means that at least 80\% of the statements need to be cited.
Subsequently, during the fine-grained preference optimization, the model is further trained for one additional epoch using a learning rate of 1e-5.

During the evaluation, we utilize the vLLM framework \citep{kwon2023efficient} for efficient inference. Without special instructions, the sampling parameters are specifically configured with a temperature of 1.0 and a top-p setting of 0.95. We present detailed prompts used during the evaluation process in Figure~\ref{fig:prompt_ablations}.
\end{document}